\documentclass[10pt,twocolumn,letterpaper]{article}

\usepackage{cvpr}
\usepackage{times}
\usepackage{epsfig}
\usepackage{graphicx}
\usepackage{amsmath}
\usepackage{amssymb}
\usepackage{bm}
\usepackage{algorithm}
\usepackage{algorithmic}

\usepackage[pagebackref=true,breaklinks=true,letterpaper=true,colorlinks,bookmarks=false]{hyperref}

\cvprfinalcopy 

\def\cvprPaperID{2762} 

\ifcvprfinal\fi
\begin{document}
	
	\title{CNN-SLAM: Real-time dense monocular SLAM with learned depth prediction}
	
	\author{Keisuke Tateno\thanks{The first two authors contribute equally to this paper.}	$^{1,2}$,  Federico Tombari$^{\ast1}$,  Iro Laina$^{1}$, Nassir Navab$^{1,3}$\\	
		\tt\small \{tateno, tombari, laina, navab\}@in.tum.de \\
		$^{1}$ CAMP - TU Munich \hspace{1.5cm} 
		$^{2}$ Canon Inc.\hspace{1.2cm}
		$^{3}$ Johns Hopkins University\\
		\hspace{-0.8cm} Munich, Germany \hspace{1.8cm} Tokyo, Japan \hspace{2.1cm} Baltimore, US
	}

	\maketitle
	
	\begin{abstract}
		Given the recent advances in depth prediction from Convolutional Neural Networks (CNNs), this paper investigates how predicted depth maps from a deep neural network can be deployed for accurate and dense monocular reconstruction. We propose a method where CNN-predicted dense depth maps are naturally fused together with depth measurements obtained from direct monocular SLAM. Our fusion scheme privileges depth prediction in image locations where monocular SLAM approaches tend to fail, e.g. along low-textured regions, and vice-versa. We demonstrate the use of depth prediction for estimating the absolute scale of the reconstruction, hence overcoming one of the major limitations of monocular SLAM. Finally, we propose a framework to efficiently fuse semantic labels, obtained from a single frame, with dense SLAM, yielding semantically coherent scene reconstruction from a single view. Evaluation results on two benchmark datasets show the robustness and accuracy of our approach. 
	\end{abstract}
	
	\section{Introduction}
	
	Structure-from-Motion (SfM) and Simultaneous Localization and Mapping (SLAM) are umbrella names for a highly active research area in the field of computer vision and robotics for the goal of 3D scene reconstruction and camera pose estimation from 3D and imaging sensors. Recently, real-time SLAM methods aimed at fusing together range maps obtained from a moving depth sensor have witnessed an increased popularity, since they can be employed for navigation and mapping of several types of autonomous agents, from mobile robots to drones, as well as for many augmented reality and computer graphics applications. This is the case of volumetric fusion approaches such as Kinect Fusion \cite{Newcombe2011kinect}, as well as dense SLAM methods based on RGB-D data \cite{Whelan2014, Keller2013}, which, in addition to navigation and mapping, can also be employed for accurate scene reconstruction. However, a main drawback of such approaches is that depth cameras have several limitations: indeed, most of them have a limited working range, and those based on active sensing cannot work (or perform poorly) under sunlight, thus making reconstruction and mapping less precise if not impossible in outdoor environments.  
	
	\begin{figure}[t]     
		\centering
		\includegraphics[width=0.95\linewidth]{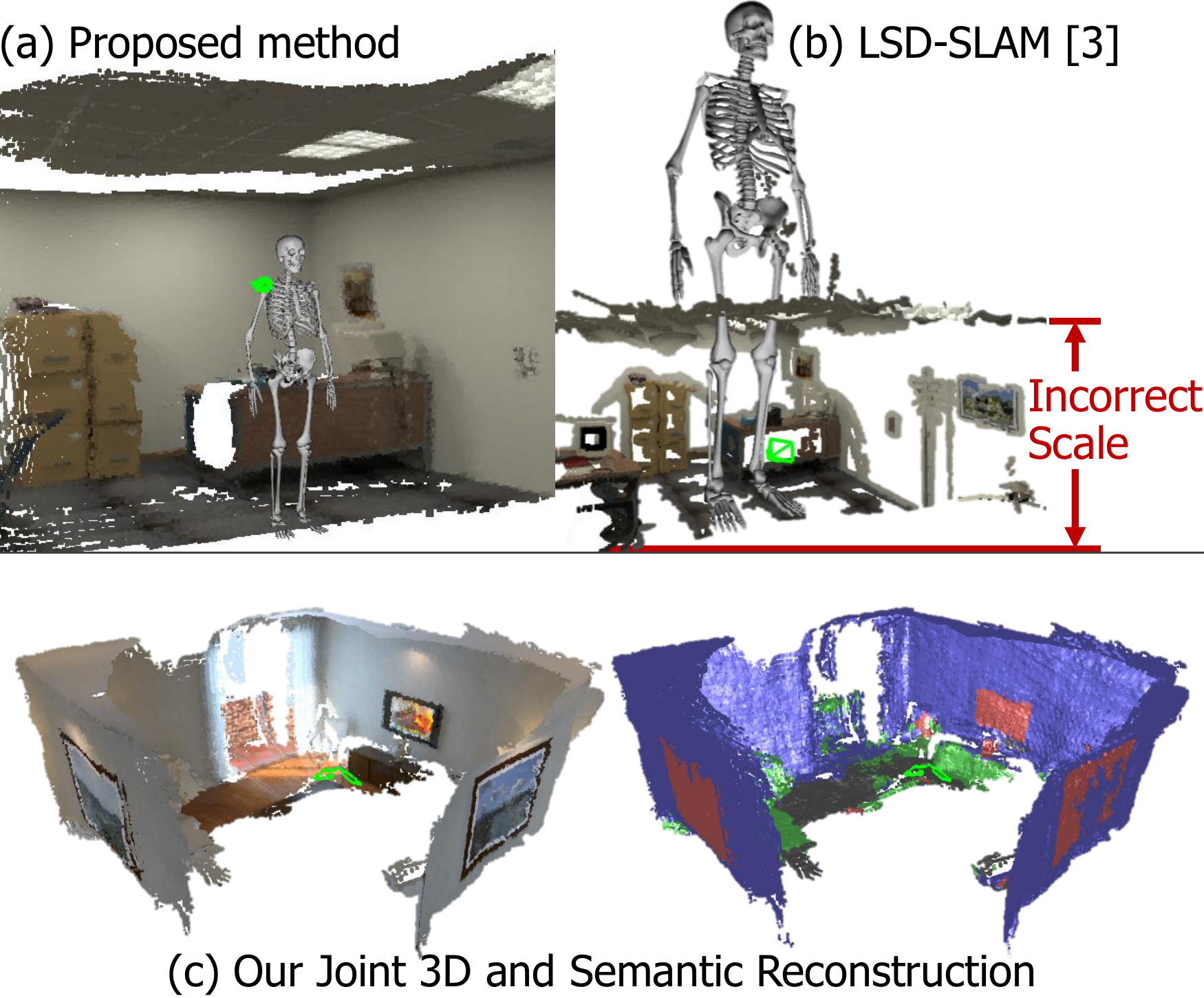}
		\caption{The proposed monocular SLAM approach (a) can estimate a much better absolute scale than the state of the art (b), which is necessary for many SLAM applications such as AR, \eg the skeleton is augmented into the reconstruction.  c) our approach can yield joint 3D and semantic reconstruction from a single view.}
		\label{fig:teaser}
		\vspace{-0.3cm}
	\end{figure}   
	
	In general, since depth cameras are not as ubiquitous as color cameras, a lot of research interest has been focused on dense and semi-dense SLAM methods from a single camera \cite{Newcombe2011dtam, Engel2014, MurArtal2015}. These approaches aim at real-time monocular scene reconstruction by estimating the depth map of the current viewpoint through small-baseline stereo matching over pairs of nearby frames. The working assumption is that the camera translates in space over time, so that pairs of consecutive frames can be treated as composing a stereo rig. Stereo matching is usually carried out through color consistency or by relying on keypoint extraction and matching. 
	
	One main limitation of monocular SLAM approaches is the estimation of the absolute scale. Indeed, even if camera pose estimation and scene reconstruction are carried out accurately, the absolute scale of such reconstruction remains inherently ambiguous, limiting the use of monocular SLAM within most aforementioned applications in the field of augmented reality and robotics (an example is shown in Fig. \ref{fig:teaser},b). Some approaches suggest solving the issue via object detection by matching the scene with a pre-defined set of 3D models, so to recover the initial scale based on the estimated object size \cite{Galvez-Lopez2016}, which nevertheless fails in absence of known shapes in the scene. Another main limitation of monocular SLAM is represented by pose estimation under pure rotational camera motion, in which case stereo estimation cannot be applied due to the lack of a stereo baseline, resulting in tracking failures. 
	
	Recently, a new avenue of research has emerged that addresses depth prediction from a single image by means of learned approaches. In particular, the use of deep Convolutional Neural Networks (CNNs) \cite{Laina2016, Eigen2014, Eigen2015} in an end-to-end fashion has demonstrated the potential of regressing depth maps at a relatively high resolution and with a good absolute accuracy even under the absence of monocular cues (texture, repetitive patterns) to drive the depth estimation task. 
	One advantage of deep learning approaches is that the absolute scale can be learned from examples and thus predicted from a single image without the need of scene-based assumptions or geometric constraints, unlike \cite{Hoiem2005, Liu2010, Delage2006}. A major limitation of such depth maps is the fact that, although globally accurate, depth borders tend to be locally blurred: hence, if such depths are fused together for scene reconstruction as in \cite{Laina2016}, the reconstructed scene will overall lack shape details. 
	
	
	Relevantly, despite the few methods proposed for single view depth prediction, the application of depth prediction to higher-level computer vision tasks has been mostly overlooked so far, with just a few examples existing in literature \cite{Laina2016}. The main idea behind this work is to exploit the best from both worlds and propose a monocular SLAM approach that fuses together depth prediction via deep networks and direct monocular depth estimation so to yield a dense scene reconstruction that is at the same time unambiguous in terms of absolute scale and robust in terms of tracking. 
	To recover blurred depth borders, the CNN-predicted depth map is used as initial guess for dense reconstruction and successively refined by means of a direct SLAM scheme relying on small-baseline stereo matching similar to the one in \cite{Engel2014}. 
	Importantly, small-baseline stereo matching holds the potential to refine edge regions on the predicted depth image, which is where they tend to be more blurred. At the same time, the initial guess obtained from the CNN-predicted depth map can provide absolute scale information to drive pose estimation, so that the estimated pose trajectory and scene reconstruction can be significantly more accurate in terms of absolute scale compared to the state of the art in monocular SLAM. Fig.~\ref{fig:teaser}, a) shows an example illustrating the usefulness of carrying out scene reconstruction with a precise absolute scale such as the one proposed in this work. Moreover, tracking can be made more robust, as the CNN-predicted depth does not suffer from the aforementioned problem of pure rotations, as it is estimated on each frame individually. Last but not least, this framework can run in real-time since the two processes of depth prediction from CNNs and depth refinement can be simultaneously carried out on different computational resources of the same architecture - respectively, the GPU and the CPU. 
	
	Another relevant aspect of recent CNNs is that the same network architecture can be successfully employed for different high-dimensional regression tasks rather than just depth estimation: one typical example is semantic segmentation \cite{Eigen2015,wang2015}. We leverage this aspect to propose an extension of our framework that uses pixel-wise labels to coherently and efficiently fuse semantic labels with dense SLAM, so to attain semantically coherent scene reconstruction from a single view: an example is shown in Fig.~\ref{fig:teaser}, c). Notably, to the best of our knowledge semantic reconstruction has been shown only recently and only based on stereo \cite{Vineet2015} or RGB-D data \cite{Lai2014}, \ie never in the monocular case. 
	
	We validate our method with a comparison on two public SLAM benchmarks against the state of the art in monocular SLAM and depth estimation, focusing on the accuracy of pose estimation and reconstruction. Since the CNN-predicted depth relies on a training procedure, we show experiments where the training set is taken from a completely different environment and a different RGB sensor than those available in the evaluated benchmarks, so to portray the capacity of our approach - particularly relevant for practical uses - to generalize to novel, unseen environments. We also show qualitative results of our joint scene reconstruction and semantic label fusion in a real environment.

	\begin{figure*}[t]     
		\centering
		\includegraphics[width=0.8\linewidth]{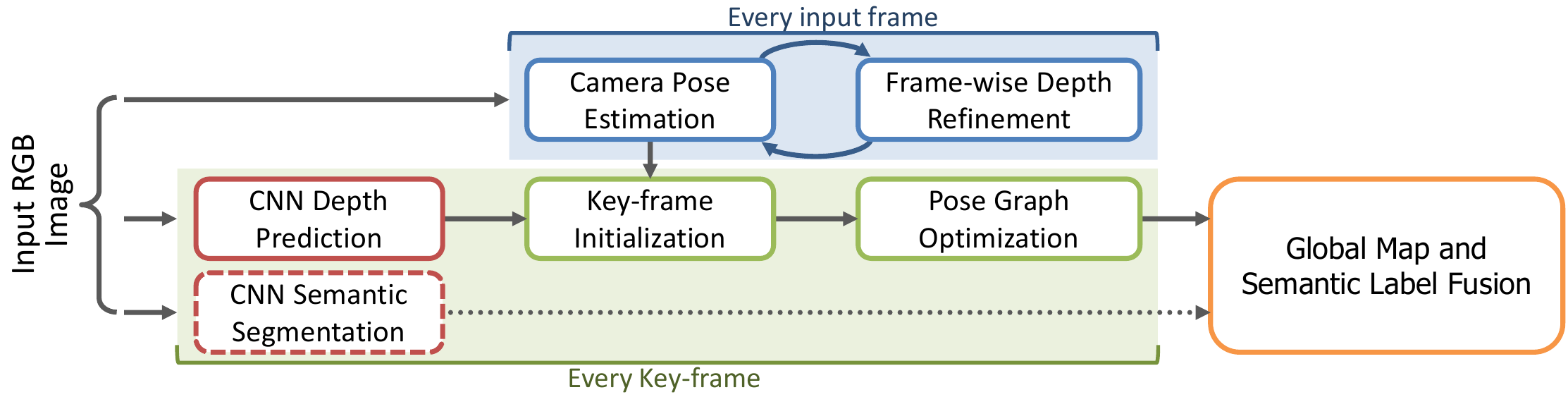}
		\caption{CNN-SLAM Overview.}
		\label{fig:system}
		\vspace{-0.3cm}
	\end{figure*} 
	
	\section{Related work} 
	In this Section we review related work with respect to the two fields that we integrate within our framework, \ie SLAM and depth prediction. 
	
	\paragraph{SLAM}There exists a vast literature on SLAM. From the point of view of the type of input data being processed, approaches can be classified into either depth camera-based \cite{Newcombe2011kinect, Whelan2014, Keller2013} and monocular camera-based \cite{Newcombe2011dtam, Engel2014, MurArtal2015}. Instead, from a methodological viewpoint, they are classified as either feature-based \cite{Klein2007, Klein2008, MurArtal2015} and direct \cite{Newcombe2011dtam, Engel2013, Engel2014}. 
	Given the scope of this paper, we will focus here only on monocular SLAM approaches.   
	
	As for feature-based monocular SLAM, ORB-SLAM \cite{MurArtal2015} is arguably the state of the art in terms of pose estimation accuracy. This method relies on the extraction of sparse ORB features from the input image to carry out a sparse reconstruction of the scene as well as to estimate the camera pose, also employing local bundle adjustment and pose graph optimization. 
	As for direct monocular SLAM, the Dense Tracking and Mapping (DTAM) of \cite{Newcombe2011dtam} achieved dense reconstruction in real-time on s GPU by using short-baseline multiple-view stereo matching with a regularization scheme, so that depth estimation is smoother on low-textured regions in the color image. Moreover, the Large-Scale Direct SLAM (LSD-SLAM) algorithm \cite{Engel2014} proposed the use of a semi-dense map representation which keeps track of depth values only on gradient areas of the input image, this allowing enough efficiency to enable direct SLAM in real-time on a CPU. An extension of LSD-SLAM is the recent Multi-level mapping (MLM) algorithm \cite{Greene2016}, which proposed the use of a dense approach on top of LSD-SLAM in order to increase its density and improve the reconstruction accuracy. 
	
	\paragraph{Depth prediction from single view}
	Depth prediction from single view has gained increasing attention in the computer vision community thanks to the recent advances in deep learning. Classic depth prediction approaches employ hand-crafted features and probabilistic graphical models \cite{Hoiem2005, Liu2010} to yield regularized depth maps, usually making strong assumptions on the scene geometry. Recently developed deep convolutional architectures significantly outperformed previous methods in terms of depth estimation accuracy \cite{Laina2016, Eigen2014, Eigen2015, wang2015, Liu15, Li15}. Interestingly, the work of \cite{Laina2016} reports qualitative results of employing depth predictions for dense SLAM as an application. In particular, the predicted depth map is used as input for Keller's Point-Based Fusion RGB-D SLAM algorithm \cite{Keller2013}, showing that SLAM-based scene reconstruction can be obtained using depth prediction, although it lacks shape details, mostly due to the aforementioned blurring artifacts that are associated with the loss of fine spatial information through the contractive part of a CNN.

	\section{Proposed Monocular Semantic SLAM}
	In this section, we illustrate the proposed framework for 3D reconstruction, where CNN-predicted dense depth maps are fused together with depth measurements obtained from direct monocular SLAM. Additionally, we show how CNN-predicted semantic segmentation can also be coherently fused with the global reconstruction model. The flow diagram in Fig.~\ref{fig:system} sketches the pipeline of our framework. We employ a key-frame based SLAM paradigm \cite{Klein2007, Engel2014, MurArtal2015}, in particular we use as baseline the direct semi-dense approach in \cite{Engel2014}. Within such approach, a subset of visually distinct frames is collected as key-frames, whose pose is subject to global refinement based on pose graph optimization. At the same time, camera pose estimation is carried out at each input frame, by estimating the transformation between the frame and its nearest key-frame. 
	
	To maintain a high frame-rate, we propose to predict a depth map via CNN only on key-frames. In particular, if the currently estimated pose is far from that of existing key-frames, a new key-frame is created out of the current frame and its depth estimated via CNN. Moreover an uncertainty map is constructed by measuring the pixel-wise confidence of each depth prediction. Since in most cases the camera used for SLAM differs from the one used to acquire the dataset on which the CNN is trained, we propose a specific normalization procedure of the depth map designed to gain robustness towards different intrinsic camera parameters. When additionally carrying out semantic label fusion, we employ a second convolutional network to predict a semantic segmentation of the input frame. Finally, a pose graph on key-frames is created so to globally optimize their relative pose.
	
	A particularly important stage of the framework, also representing one main contribution of our proposal, is the scheme employed to refine the CNN-predicted depth map associated to each key-frame via small-baseline stereo matching, by enforcing color consistency minimization between a key-frame and associated input frames. In particular, depth values will be mostly refined around image regions with gradients, i.e. where epipolar matching can provide improved accuracy. This will be outlined in Subsections \ref{sec:depth_initialization} and \ref{sec:depth_refinement}. Relevantly, the way refined depths are propagated is driven by the uncertainty associated to each depth value, estimated according to a specifically proposed confidence measure (defined in Subsec. \ref{sec:depth_initialization}). 
	Every stage of the framework is now detailed in the following Subsections.

	\subsection{Camera Pose Estimation}
	\label{sec:pose_estimation}
	
	The camera pose estimation is inspired by the key-frame approach in \cite{Engel2014}. In particular, the system holds a set of key-frames $k_1, .., k_n \in \mathcal{K}$ as structural elements on which to perform SLAM reconstruction. Each key-frame $k_i$ is associated to a key-frame pose $\bm{T}_{k_i}$, a depth map $\mathcal{D}_{k_i}$, and a depth uncertainty map $\mathcal{U}_{k_i}$. In contrast to \cite{Engel2014}, our depth map is dense because it is generated via CNN-based depth prediction (see Subsec.~\ref{sec:depthprediction}). The uncertainty map measures the confidence of each depth value. As opposed to \cite{Engel2014} that initializes the uncertainty to a large, constant value, our approach initializes it according to the measured confidence of the depth prediction (described in Subsec.~\ref{sec:depth_initialization}). 
	In the following, we will refer to a generic depth map element as $\bm{u} = (x,y)$, which ranges in the image domain, i.e. $\bm{u} \in \Omega \subset \mathbb{R}^2$, with $\dot{\bm{u}}$ being its homogeneous representation. 
	
	
	At each frame $t$, we aim to estimate the current camera pose $\bm{T}^{k_i}_t = [\bm{R}_t,\bm{t}_t] \in \mathbb{SE}(3)$, i.e. the transformation between the nearest key-frame $k_i$ and frame $t$, composed of a 3$\times$3 rotation matrix $\bm{R}_t \in \mathbb{SO}(3)$ and a 3D translation vector $\bm{t}_t \in \mathbb{R}^3$. This transformation is estimated by minimizing the photometric residual between the intensity image $\mathcal{I}_t$ of the current frame and the intensity image $\mathcal{I}_{k_i}$ 
	of the nearest key-frame $k_i$, via weighted Gauss-Newton optimization based on the objective function
	\begin{equation} 
	\displaystyle E (\bm{T}^{k_i}_t) =\sum_{\tilde{\bm{u}}  \in \Omega } \rho \left( \frac{ r \left(  \tilde{\bm{u}} , \bm{T}^{k_i}_t \right)  }{ \sigma \left( r \left( \tilde{\bm{u}} , \bm{T}^{k_i}_t \right)  \right) } \right)
	\end{equation}
	where $\rho$ is the Huber norm and $\sigma$ is a function measuring the residual uncertainty \cite{Engel2014}. Here, $r$ is the photometric residual defined as 
	\begin{equation} 
	\label{eq_residual}
	r \left( \tilde{\bm{u}} , \bm{T}^{k_i}_t \right)  = \mathcal{I}_{k_i}( \tilde{\bm{u}}  ) -  \mathcal{I}_t\left( \pi \left(\bm{K} \bm{T}^{k_i}_t \tilde{\mathcal{V}}_{k_i} \left( \tilde{\bm{u}}  \right) \right) \right) ~.
	\end{equation}
	
	Considering that our depth map is dense, for the sake of efficiency, we limit the computation of the photometric residual only on the subset of pixels lying within high color gradient regions, defined by the image domain subset $\tilde{\bm{u}} \subset \bm{u} \in \Omega $. Also, in (\ref{eq_residual}), $\pi$ represents the perspective projection function mapping a 3D point to a 2D image coordinate
	\begin{equation} 
	\pi\left(  [x y z]^T \right) = \left( x / z, y / z \right)^T
	\end{equation} 
	while $\mathcal{V}_{k_i}( \bm{u} )$ represents a 3D element of the vertex map computed from the key-frame's depth map
	\begin{equation} 
	\mathcal{V}_{k_i}( \bm{u} )=\bm{K}^{-1} \dot{\bm{u} } \mathcal{D}_{k_i}\left( \bm{u} \right)
	\end{equation} 
	where $\bm{K}$ is the camera intrinsic matrix. 
	
	Once $\bm{T}^{k_i}_t$ is obtained, the current camera pose in world coordinate system is computed as $\bm{T}_t = \bm{T}^{k_i}_t \bm{T}_{k_i}$. 

	\subsection{CNN-based Depth Prediction and Semantic Segmentation}
	\label{sec:depthprediction}
	Every time a new key-frame is created, an associated depth map is predicted via CNN. 
	The depth prediction architecture that we employ is the state-of-the-art approach proposed in \cite{Laina2016}, based on the extension of the Residual Network (ResNet) architecture \cite{he2016} to a Fully Convolutional network. In particular, the first part of the architecture is based on ResNet-50~\cite{he2016} and initialized with pre-trained weights on ImageNet \cite{ILSVRC15}. 
	The second part of the architecture replaces the last pooling and fully connected layers originally presented in ResNet-50 with a sequence of residual up-sampling blocks composed of a combination of unpooling and convolutional layers. After up-sampling, drop-out is applied before a final convolutional layer which outputs a 1-channel output map representing the predicted depth map. The loss function is based on the reverse Huber function \cite{Laina2016}. 
	
	Following the successful paradigm of other approaches that employed the same architecture for both depth prediction and semantic segmentation tasks \cite{Eigen2015,wang2015}, we also re-trained this network for predicting pixel-wise semantic labels from RGB images. To deal with this task, we modified the network so that it has as many output channels as the number of categories and employed a soft-max layer and a cross-entropy loss function to be minimized via back-propagation and Stochastic Gradient Descent (SGD).
	It is important to point out that, although in principle any semantic segmentation algorithm could be used, the primary objective of this work is to showcase how frame-wise segmentation maps can be successfully fused within our monocular SLAM framework (see Subsec.~\ref{sec:label_fusion}).

	\subsection{Key-frame Creation and Pose Graph Optimization}
	\label{sec:depth_initialization}
	
	
	One limitation of using a pre-trained CNN for depth prediction is that, if the sensor used for SLAM has different intrinsic parameters from those used to capture the training set, the resulting absolute scale of the 3D reconstruction will be inaccurate. To ameliorate this issue, we propose to adjust the depth regressed via CNN with the ratio between the focal length of the current camera, $f_{cur}$ and that of the sensor used for training, $f_{tr}$ as
	\begin{equation}
	\label{eq_adjustment}
	\mathcal{D}_{k_i} \left( \bm{u} \right)  =  \frac{ f_{cur} } { f_{tr}} \tilde{\mathcal{D}_{k_i}} \left( \bm{u} \right)
	\end{equation}
	where $\tilde{\mathcal{D}_{k_i}}$ is the depth map directly regressed by the CNN from the current key-frame image $\mathcal{I}_i$.
	
\begin{figure}[t]     
		\centering
		\includegraphics[width=\linewidth]{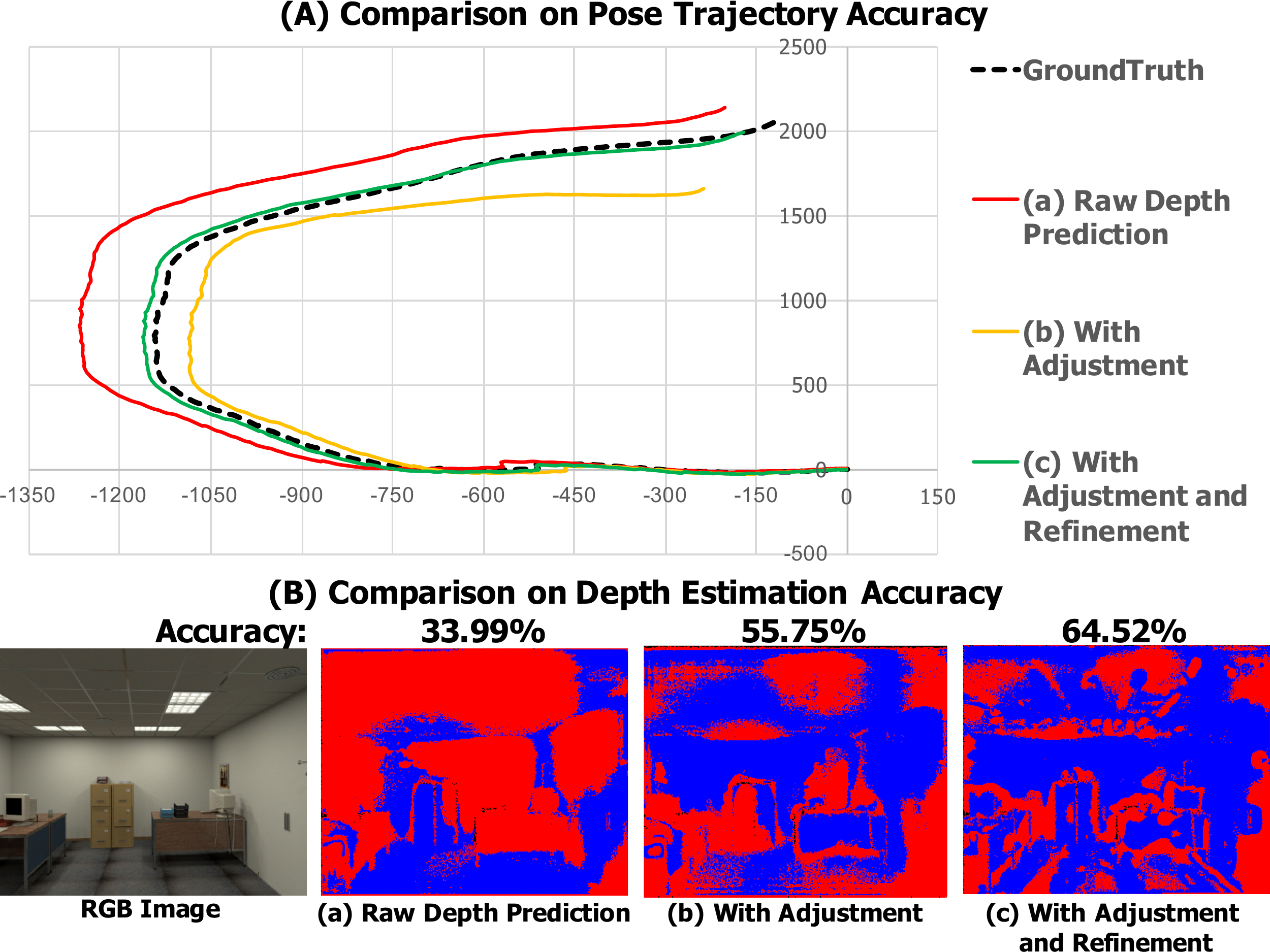}
		\caption{ Comparison among (a) direct CNN-depth prediction, (b) after depth adjustment and (c) after depth adjustment and refinement, in terms of (A) pose trajectory accuracy and (B) depth estimation accuracy. Blue pixels depict correctly estimated depths, i.e. within 10 \% of ground-truth. The comparison is done on one sequence of the \emph{ICL-NUIM} dataset \cite{Handa2014}.}
		\label{fig:effectiveness}
		\vspace{-0.1cm}
	\end{figure}

	Fig. \ref{fig:effectiveness} shows the usefulness of the adjustment procedure defined in (\ref{eq_adjustment}), on a sequence of the benchmark \emph{ICL-NUIM} dataset \cite{Handa2014} (compare \emph{(a)} with \emph{(b)} ). As shown, the performance after the adjustment procedure is significantly improved over that of using the depth map as directly predicted by the CNN. The improvement shows both in terms of depth  accuracy as well as pose trajectory accuracy. 
	
	In addition, we associate each depth map $\mathcal{D}_{k_i}$ to an uncertainty map $\mathcal{U}_{k_i}$. In \cite{Engel2014}, this map is initialized by setting each element to a large, constant value. Since the CNN provides us with dense maps at each frame but without relying on any temporal regularization, we propose to instead initialize our uncertainty map by computing a confidence value based on the difference between the current depth map and its respective scene point on the nearest key-frame. Thus, this confidence measures how coherent each predicted depth value is across different frames: for those elements associated to a high confidence, the successive refinement process will be much faster and effective than the one in \cite{Engel2014}. 
	
	Specifically, the uncertainty map $\mathcal{U}_{k_i}$ is defined as the element-wise squared difference between the depth map of the current key-frame $k_i$ and that of the nearest key-frame $k_j$, warped according to the estimated transformation $\bm{T}^{k_i}_{k_j}$ from $k_i$ to $k_j$  
	\begin{equation} 
	\mathcal{U}_{k_i} \left( \bm{u}  \right) = \left( \mathcal{D}_{k_i} \left( \bm{u} \right) - \mathcal{D}_{k_j} \left( \pi \left( \bm{K} \bm{T}^{k_i}_{k_j} \mathcal{V}_{k_i} \left( \bm{u} \right) \right) \right) \right)^2 .
	\end{equation}
	
	
	To further improve the accuracy of each newly initialized key-frame, we propose to fuse its depth map and uncertainty map with those propagated from the nearest key-frame (this will obviously not apply to the very first key-frame) after they have been refined with new input frames (the depth refinement process is described in Subsection \ref{sec:depth_refinement}). To achieve this goal, we first define a propagated uncertainty map from the nearest key-frame $k_j$ as
	\begin{equation}
	\tilde{\mathcal{U}}_{k_j} \left( \bm{v} \right) = \frac{D_{k_j} \left( \bm{v} \right)}{D_{k_i} \left( \bm{u} \right)} \mathcal{U}_{k_j} \left( \bm{v} \right) + \sigma^{2}_p
	\end{equation}
	where $\bm{v} = \pi \left( \bm{K} \bm{T}^{k_i}_{k_j} \mathcal{V}_{k_i} \left( \bm{u} \right) \right)$ while, following \cite{Engel2014}, $\sigma^{2}_p$ is the white noise variance used to increase the propagated uncertainty. 
	Then, the two depth maps and uncertainty maps are fused together according to the weighted scheme
	\begin{eqnarray} 
	\label{eq:fusion_keyframes}
	& \mathcal{D}_{k_i} \left( \bm{u} \right)  =  \frac{ \tilde{\mathcal{U}}_{k_j} \left( \bm{v} \right) \cdot \mathcal{D}_{k_i} \left( \bm{u} \right) + \mathcal{U}_{k_i}\left( \bm{u} \right) \cdot \mathcal{D}_{k_j} \left( \bm{v} \right)} { \mathcal{U}_{k_i}\left( \bm{u} \right) + \tilde{\mathcal{U}}_{k_j} \left( \bm{v} \right) } \\
	& \mathcal{U}_{k_i} \left( \bm{u} \right) = \frac{ \tilde{\mathcal{U}}_{k_j} \left( \bm{v} \right) \cdot \mathcal{U}_{k_i} \left( \bm{u} \right) } { \mathcal{U}_{k_i} \left( \bm{u} \right) + \tilde{\mathcal{U}}_{k_j}\left( \bm{v} \right) } ~.
	\end{eqnarray}
	

	Finally, the pose graph is also updated at each new key-frame, by creating new edges with the key-frames already present in the graph that share a similar field of view (i.e., having a small relative pose) with the newly added key-frame. Moreover, the pose of the key-frames is each time globally refined via pose graph optimization \cite{Kuemmerle2011}. 
	

	\subsection{Frame-wise Depth Refinement}
	\label{sec:depth_refinement}
	
	The goal of this stage is to continuously refine the depth map of the currently active key-frame based on the depth maps estimated at each new frame. To achieve this goal, we use the small baseline stereo matching strategy described in the semi-dense scheme of \cite{Engel2013}, by computing at each pixel of the current frame $t$ a depth map $\mathcal{D}_t$ and an uncertainty map $\mathcal{U}_t$ based on the 5-pixel matching along the epipolar line. These two maps are aligned with the key-frame $k_i$ based on the estimated camera pose $\bm{T}^{k_i}_t$.
	
	The estimated depth map and uncertainty map are then directly fused with those of the nearest key-frame $k_i$ as follows:
	\begin{eqnarray} 
	\label{eq:fusion}
	& \mathcal{D}_{k_i} \left( \bm{u} \right)  =  \frac{ \mathcal{U}_{t} \left( \bm{u} \right) \cdot \mathcal{D}_{k_i} \left( \bm{u} \right) + \mathcal{U}_{k_i}\left( \bm{u} \right) \cdot \mathcal{D}_{t} \left( \bm{u} \right)} {\mathcal{U}_{k_i}\left( \bm{u} \right) + \mathcal{U}_{t}\left( \bm{u} \right) } \\
	& \mathcal{U}_{k_i} \left( \bm{u} \right) =  \frac{ \mathcal{U}_{t} \left( \bm{u} \right) \cdot \mathcal{U}_{k_i} \left( \bm{u} \right)}{\mathcal{U}_{k_i} \left( \bm{u} \right) + \mathcal{U}_{t}\left( \bm{u} \right) } 
	\end{eqnarray}
	
	Importantly, since the key-frame is associated to a dense depth map thanks to the proposed CNN-based prediction, this process can be carried out densely, i.e. every element of the key-frame is refined, in contrast to \cite{Engel2013} that only refines depth values along high gradient regions. Since the observed depths within low-textured regions tend to have a high-uncertainty (i.e., a high value in $\mathcal{U}_{t}$), the proposed approach will naturally lead to a refined depth map where elements in proximity of high intensity gradients will be refined by the depth estimated at each frame, while elements within more and more low-textured regions will gradually hold the predicted depth value from the CNN, without being affected from uncertain depth observations.

	Fig.~\ref{fig:effectiveness} demonstrates the effectiveness of the proposed depth map refinement procedure on a sequence of the benchmark \emph{ICL-NUIM} dataset \cite{Handa2014}. The Figure reports, in \emph{(c)}, the performance obtained after both adjustment and depth refinement of the depth map, showing a significant improvement of both depth estimation and pose trajectory with respect to the previous cases.

	\subsection{Global Model and Semantic Label Fusion}
	\label{sec:label_fusion}
	The obtained set of key-frames can be fused together to generate a 3D global model of the reconstructed scene. Since the CNN is trained to provide semantic labels in addition to depth maps, semantic information can be also associated to each element of the 3D global model, through a process that we denote as semantic label fusion. 
	
	In our framework, we employ the real-time scheme proposed in \cite{Tateno2015}, which aims at incrementally fusing together the depth map and the connected component map obtained from each frame of a RGB-D sequence. This approach uses a Global Segmentation Model (GSM) to average the assignment  of labels to each 3D element over time, so to be robust to noise in the frame-wise segmentation. 
	In our case, the pose estimation is provided as input to the algorithm, since camera poses are estimated via monocular SLAM, while input depth maps are those associated to the set of collected key-frames only. Here, instead of connected component maps as in \cite{Tateno2015}, we use semantic segmentation maps. The result is a 3D reconstruction of the scene, incrementally built over new key-frames, where each 3D element is associated to a semantic class from the set used to train the CNN. 
	

	\section{Evaluation}
	\label{sec:evaluation}
	We provide here an experimental evaluation to validate the contributions of our method in terms of tracking and reconstruction accuracy, by means of a quantitative comparison against the state of the art on two public benchmark datasets (Subsec. \ref{sec:eval_quantitative}), 
	as well as a qualitative assessment in terms of robustness against pure rotational camera motions (Subsec. \ref{sec:eval_rotation}) and accuracy of semantic label fusion (Subsec. \ref{sec:eval_semantic}). 
	
	The evaluation is carried out on a desktop PC with an Intel Xeon CPU at 2.4GHz with 16GB of RAM and a Nvidia Quadro K5200 GPU with 8GB of VRAM. As for the implementation of our method, although the CNN network works on an input/output resolution of 304$\times$228 \cite{Laina2016}, both the input frame and the predicted depth map are converted to 320$\times$240 as input for all other stages. Also, the CNN-based depth prediction and semantic segmentation are run on the GPU, while all other stages are implemented on the CPU, and run on two different CPU threads, one devoted to frame-wise processing stages (camera pose estimation and depth refinement), the other carrying out key-frame related processing stages (key-frame initialization, pose graph optimization and global map and semantic label fusion), so to allow our entire framework to run in real-time. 

	\begin{table*}[t]
		\caption{Comparison in terms of Absolute Trajectory Error [m] and percentage of correctly estimated depth on ICL-NUIM and TUM datasets (TUM/seq1: \emph{fr3/long\_office\_household}, TUM/seq2: \emph{fr3/nostructure\_texture\_near\_withloop}, TUM/seq3: \emph{fr3/structure\_texture\_far}.}
		\label{table:eval_ate} \vspace{-0.2cm}
		\begin{center}
			\scalebox{0.88}{
				\begin{tabular}{|c||c|c|c|c|c||c|c|c|c|c|c|}
					\hline
					& \multicolumn{5}{|c||}{\textbf{Abs. Trajectory Error}} & \multicolumn{6}{|c|}{\textbf{Perc. Correct Depth}} \\
					\hline
					& \multicolumn{1}{|c|}{Our} & \multicolumn{1}{|c|}{LSD-BS} & \multicolumn{1}{|c|}{LSD} & \multicolumn{1}{|c|}{ORB} & \multicolumn{1}{|c||}{Laina } & \multicolumn{1}{|c|}{Our} & \multicolumn{1}{|c|}{LSD-BS} & \multicolumn{1}{|c|}{LSD} & \multicolumn{1}{|c|}{ORB} & \multicolumn{1}{|c|}{Laina} & \multicolumn{1}{|c|}{Remode} \\              
					& \multicolumn{1}{|c|}{Method} & \multicolumn{1}{|c|}{\cite{Engel2014}} & \multicolumn{1}{|c|}{\cite{Engel2014}} & \multicolumn{1}{|c|}{\cite{MurArtal2015}} & \multicolumn{1}{|c||}{\cite{Laina2016}} & \multicolumn{1}{|c|}{Method} & \multicolumn{1}{|c|}{\cite{Engel2014}} & \multicolumn{1}{|c|}{\cite{Engel2014}} & \multicolumn{1}{|c|}{\cite{MurArtal2015}} & \multicolumn{1}{|c|}{\cite{Laina2016} } & \multicolumn{1}{|c|}{\cite{Pizzoli2014}} \\
					\hline
					\hline
					ICL/office0& $\bm{0.266}$  & 0.587 & 0.528 & 0.430 & 0.337 & $\bm{19.410}$ & 0.603 & 0.335 & 0.018 & 17.194  & 4.479 \\
					\hline
					ICL/office1& $\bm{0.157}$ & 0.790 & 0.768 & 0.780 & 0.218 & $\bm{29.150}$ & 4.759 & 0.038 & 0.023 & 20.838 & 3.132 \\				
					\hline
					ICL/office2& 0.213 & $\bm{0.172}$ & 0.794 & 0.860 & 0.509 & $\bm{37.226}$  & 1.435 & 0.078 & 0.040 & 30.639 & 16.7081\\
					\hline
					ICL/living0 & $\bm{0.196}$ & 0.894 & 0.516 & 0.493  & 0.230 & 12.840 & 1.443 & 0.360  & 0.027  & $\bm{15.008}$  & 4.479 \\				
					\hline
					ICL/living1 & $\bm{0.059}$ & 0.540 & 0.480 & 0.129  & 0.060 & $\bm{13.038}$ & 3.030 & 0.057  & 0.021  & 11.449 & 2.427\\				
					\hline
					ICL/living2 & 0.323 & $\bm{0.211}$ & 0.667 & 0.663  & 0.380 & 26.560& 1.807 & 0.167 & 0.014  & $\bm{33.010}$ & 8.681 \\				
					\hline
					TUM/seq1 & $\bm{0.542}$ & 1.717 & 1.826  & 1.206 & 0.809 & 12.477  & 3.797  & 0.086  & 0.031 & $\bm{12.982}$ & 9.548\\
					\hline
					TUM/seq2 & 0.243  & $\bm{0.106}$  & 0.436  & 0.495  & 1.337 & $\bm{24.077}$  & 3.966  & 0.882  & 0.059  & 15.412  & 12.651\\
					\hline
					TUM/seq3 & 0.214   & $\bm{0.037}$  & 0.937  & 0.733  & 0.724 & $\bm{27.396}$  & 6.449   & 0.035   & 0.027   & 9.450 & 6.739 \\				
					\hline \hline
					\textbf{Avg.} & $\bm{0.246}$ & 0.562 & 0.772 & 0.643 & 0.512 & $\bm{22.464}$  & 3.032 & 0.226 & 0.029 & 18.452 & 7.649 \\
					\hline
				\end{tabular}
			}
		\end{center}
	\end{table*}
	
	\begin{figure*}[t]     
		\centering
		\vspace{-0.6cm}
		\includegraphics[width=.9\linewidth]{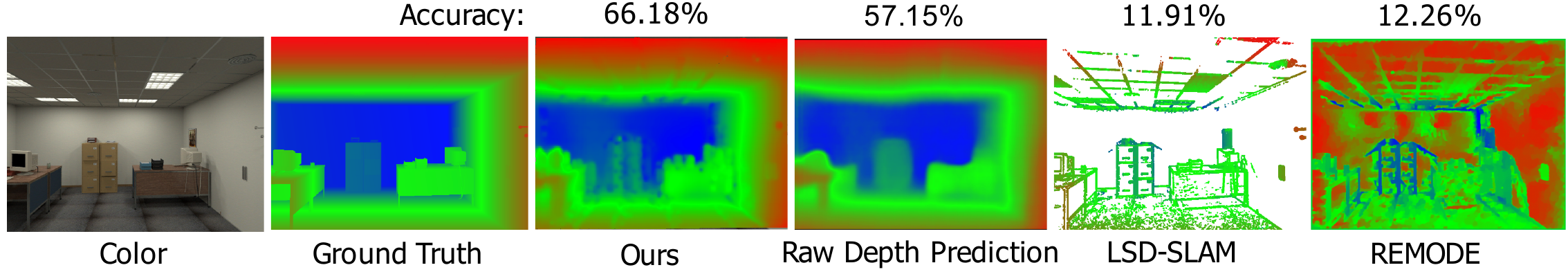}
		\caption{Comparison in terms of depth map accuracy and density among (from the left) the ground-truth, a refined key-frame from our approach, the corresponding raw depth prediction from the CNN, the refined key-frame from LSD-SLAM \cite{Engel2014} with bootstrapping and estimated dense depth map from REMODE \cite{Pizzoli2014}, on the (\emph{office2}) sequence from the \emph{ICL-NUIM} dataset \cite{Handa2014}. The accuracy value means correctly estimated depth density on this key-frame.}
		\label{fig:eval_density}
	\end{figure*} 
	
	\begin{figure*}[t]     
		\centering
		\includegraphics[width=.75\linewidth]{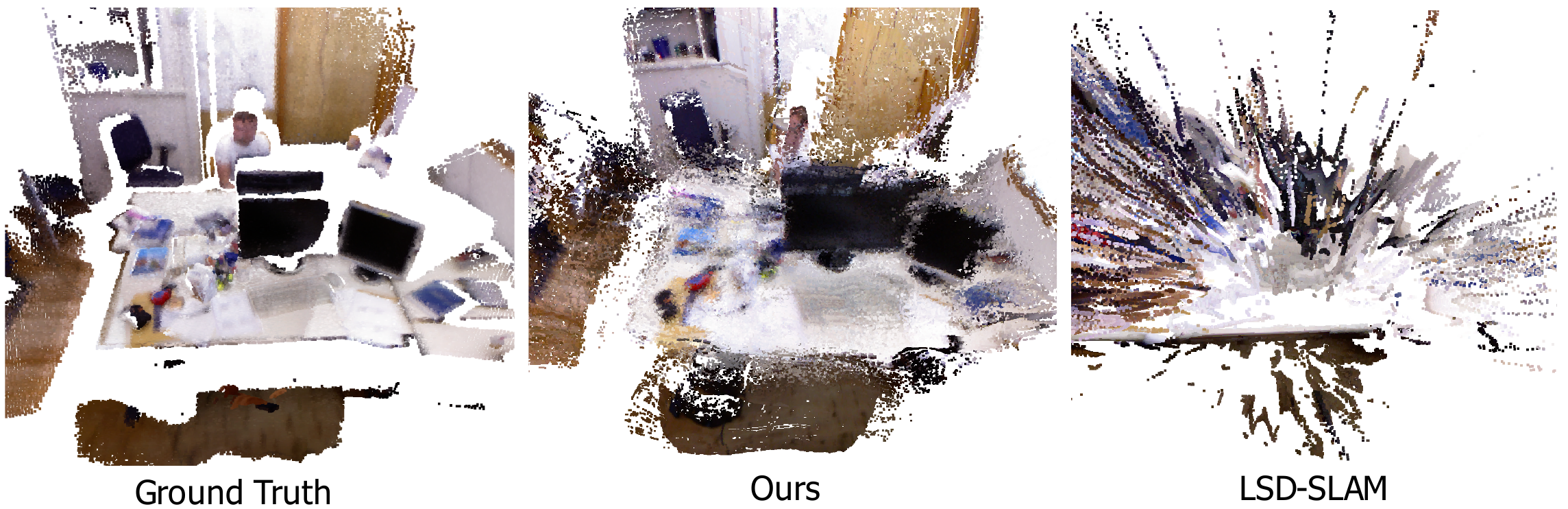}
		\caption{Comparison on a sequence that includes mostly pure rotational camera motion between the reconstruction obtained by ground truth depth (left), proposed method (middle) and LSD-SLAM \cite{Engel2014} (right). }
		\label{fig:reconst_pureRotation}
	\end{figure*} 
	
	We use sequences from two public benchmark datasets, i.e. the \emph{ICL-NUIM} dataset \cite{Handa2014} and \emph{TUM RGB-D SLAM} dataset \cite{Sturm2012}, the former synthetic, the latter acquired with a Kinect sensor. Both datasets provide ground truth in terms of camera trajectory and depth maps. In all our experiments, we used the CNN model trained on the indoor sequences of the \emph{NYU Depth v2} dataset \cite{Silberman2012}, to test the generalization capability of the network to unseen environments; also because this dataset includes both depth ground-truth (represented by depth maps acquired with a Microsoft Kinect camera) and pixel-wise semantic label annotations, necessary for semantic label fusion. 
	In particular, we train the semantic segmentation network on the official train split of the labeled subset, while the depth network is trained using more frames from the raw NYU dataset, as reported in \cite{Laina2016}. Semantic annotations consist of the 4 super-classes \emph{floor}, \emph{vertical structure}, \emph{large structure/furniture}, \emph{small structure}. Noteworthy, the settings of the training dataset are quite different from those on which we evaluate our method, since they encompass different camera sensors, viewpoints and scene layouts. For example, \emph{NYU Depth v2} includes many living rooms, kitchens and bedrooms, which are missing in TUM RGB-D SLAM, being focused on office rooms with desks, objects and people.

	\subsection{Comparison against SLAM state of the art}
	\label{sec:eval_quantitative}
	We compare our approach against the publicly available implementations of LSD-SLAM\footnote{\url{github.com/tum-vision/lsd_slam}} \cite{Engel2014} and ORB-SLAM\footnote{\url{github.com/raulmur/ORB_SLAM2}} \cite{MurArtal2015}, two state-of-the-art methods in monocular SLAM representatives of, respectively, direct and feature-based methods. 
	For completeness, we also compare against REMODE \cite{Pizzoli2014}, state-of-the-art approach focused on dense monocular depth map estimation. The implementation of REMODE has been taken from the author's code\footnote{\url{https://www.github.com/uzh-rpg/rpg_open_remode}}. 
	Finally, we also compare our method to the one in \cite{Laina2016}, that uses the CNN-predicted depth maps as input for a state-of-the-art depth-based SLAM method (point-based fusion\cite{Keller2013, Tateno2015}), based on the available implementation from the authors of \cite{Tateno2015}\footnote{\url{campar.in.tum.de/view/Chair/ProjectInSeg}}. Given the ambiguity of monocular SLAM approaches to estimate absolute scale, we also evaluate LSD-SLAM by bootstrapping its initial scale using the ground-truth depth map, as done in the evaluation in \cite{Engel2014, MurArtal2015}. 
	As for REMODE, since it requires as input the camera pose estimation at each frame, we use the trajectory and key-frames estimated by LSD-SLAM with bootstrapping.  
	
	\begin{figure*}[t]     
		\centering
		\includegraphics[width=0.96\linewidth]{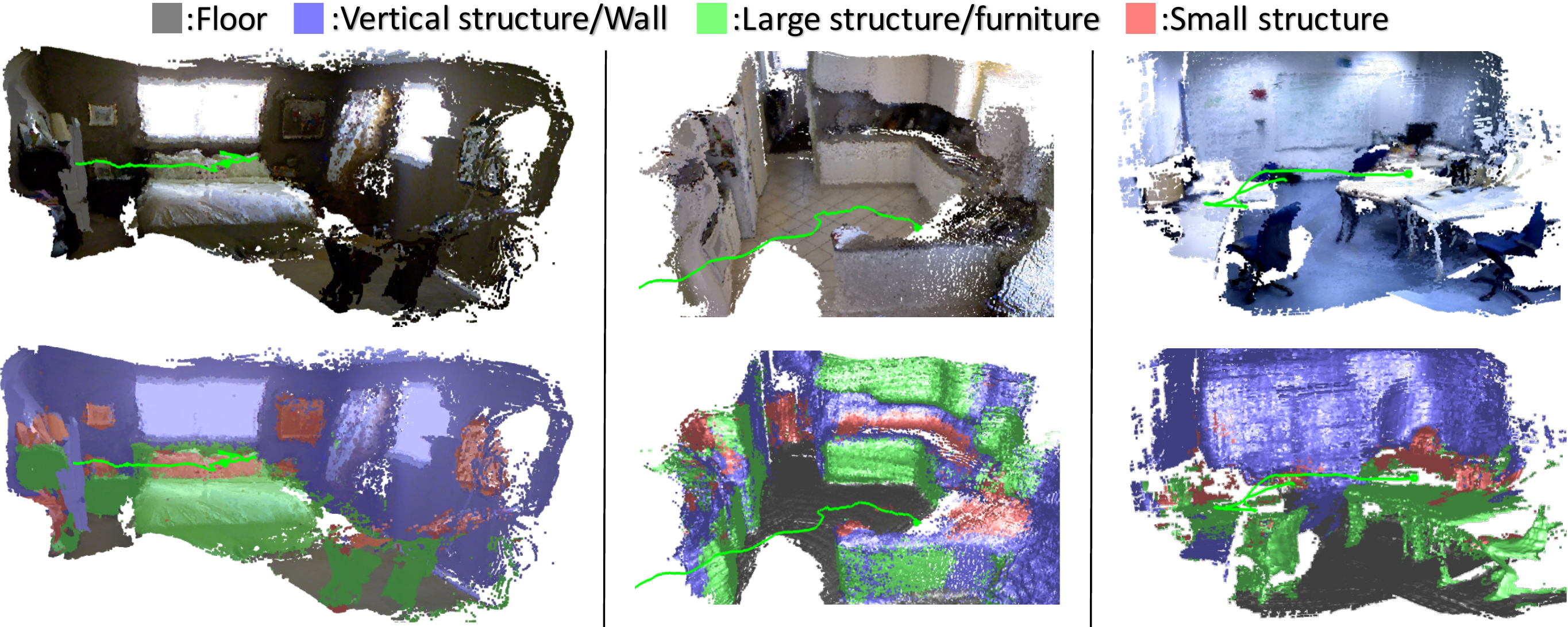}
		\caption{The results of reconstruction and semantic label fusion on the office sequence (top, acquire by our own) and one sequence (\emph{kitchen\_0046}) from the \emph{NYU Depth V2} dataset \cite{Silberman2012} (bottom). Reconstruction is shown with colors (left) and with semantic labels (right). }
		\label{fig:semanticFusion}
	\end{figure*}

	Following the evaluation methodology proposed in \cite{Sturm2012}, Table \ref{table:eval_ate} reports the camera pose accuracy based on the Absolute Trajectory Error (ATE), computed as the root mean square error between the estimated camera translation and the ground-truth camera translation for each evaluated sequence. In addition, we assess both reconstruction accuracy and density, by evaluating the percentage of depth values whose difference with the corresponding ground truth depth is less than $10\%$. Given the observations in the Table, our approach is able to always report a much higher pose trajectory accuracy with respect to monocular methods, due to the their aforementioned absolute scale ambiguity. Interestingly, the pose accuracy of our technique is on average higher than that of LSD-SLAM even after applying bootstrapping, implying an inherent effectiveness of the proposed depth fusion approach rather than just estimating the correct scaling factor. The same benefits are present in terms of reconstruction, being the estimated key-frames not only dramatically more accurate, but also much denser than those reported by LSD-SLAM and ORB-SLAM. 
	Moreover, our approach also reports a better performance in terms of both pose and reconstruction accuracy, also comparing to the technique in \cite{Laina2016}, where CNN-predicted depths are used as input for SLAM without any refinement, this again demonstrating the effectiveness of the proposed scheme to refine the blurred edges and wrongly estimated depth values predicted by the CNN. Finally, we clearly outperform also REMODE in terms of depth map accuracy. 
	
	The increased accuracy with respect to the depth maps estimated by the CNN (as employed in \cite{Laina2016}) and by REMODE, as well as the higher density with respect to LSD-SLAM is also shown in Fig. \ref{fig:eval_density}. The figure compares the ground-truth with, a refined key-frame using our approach, the corresponding raw depth prediction from the CNN, the refined key-frame from LSD-SLAM \cite{Engel2014} using bootstrapping and the estimated dense depth map from REMODE on a sequence of the \emph{ICL-NUIM} dataset. Not only our approach demonstrates a much higher density with respect to LSD-SLAM, but the refinement procedure helps to drastically reduce the blurring artifacts of the CNN-based prediction, increasing the overall depth accuracy. Also, we can note that REMODE tends to fail along low-textured regions, as opposed to our method which can estimate depth densely over such areas by leveraging the CNN-predicted depth values.

	\subsection{Accuracy under pure rotational motion}
	\label{sec:eval_rotation}
	
	As mentioned, one of the advantages of our approach compared to standard monocular SLAM is that, under pure rotational motion, the reconstruction can still be obtained by relying on CNN-predicted depths, while other methods would fail given the absence of a stereo baseline between consecutive frames. To portray this benefit, we evaluate our method on the (\emph{fr1/rpy}) sequence from the \emph{TUM} dataset, mostly consisting of just rotational camera motion. The reconstruction obtained by, respectively, our approach and LSD-SLAM compared to ground-truth are shown in Figure \ref{fig:reconst_pureRotation}. As it can be seen, our method can reconstruct the scene structure even if the camera motion is purely rotational, while the result of LSD-SLAM is significantly noisy, since the stereo baseline required to estimate depth is for most frames not sufficient. We also tried ORB-SLAM on this sequence but it completely fails, given the lack of the necessary baseline to initialize the algorithm.  
	
	\subsection{Joint 3D and semantic reconstruction}
	\label{sec:eval_semantic}
	
	Finally, we show some qualitative results of the joint 3D and semantic reconstruction achieved by our method. Three examples are shown in Fig. \ref{fig:semanticFusion}, which reports an office scene reconstructed from a sequence acquired with our own setup and two sequences from the test set of the \emph{NYU Depth V2} dataset \cite{Silberman2012}. Another example from the sequence \emph{living0} of the \emph{ICL-NUIM} dataset is shown in Fig.\ref{fig:teaser},c). The Figures also report, in green, the estimated camera trajectory. To the best of our knowledge, this is the first demonstration of joint 3D and semantic reconstruction with a monocular camera. Additional qualitative results in terms of pose and reconstruction quality as well as semantic label fusion are included in the supplementary material.
	
	\section{Conclusion}
	We have shown how the integration of SLAM with depth prediction via a deep neural network is a promising direction to solve inherent limitations of traditional monocular reconstruction, especially with respect to estimating the absolute scale, obtaining dense depths along texture-less regions and dealing with pure rotational motions. The proposed approach to refine CNN-predicted depth maps with small baseline stereo matching naturally overcomes these issues while retaining the robustness and accuracy of direct monocular SLAM in presence of camera translations and high image gradients. The overall framework is capable of jointly reconstructing the scene while fusing semantic segmentation labels with the global 3D model, opening new perspectives towards scene understanding with a monocular camera. A future research avenue is represented by closing the loop with depth prediction, i.e. improving depth estimation by means of geometrically refined depth maps.  
	
	{\small
		\bibliographystyle{ieee}

	}

\end{document}